\documentclass[11pt]{article}

\usepackage{agenticlearning-xelatex}
\usepackage{placeins}

\newcommand{\ourmethod}{{\it Context Tuning}}
\newcommand{\ourmethodprompt}{{\it CT-Prompt}}
\newcommand{\ourmethodkv}{{\it CT-KV}}
\newcommand{\ourmethodv}{{\it CT-V}}
\newcommand{\ourmethodprefix}{{\it CT-Prefix}}
\newcommand{\ourframework}{{\it In-Context Optimization}}
\newcommand{\tttct}{TTT+\ourmethodkv{}}
\newcommand{\ico}{{\it ICO}}

\begin{document}

\shorttitle{Context Tuning for In-Context Optimization}
\shortauthor{Lu \etal}

\title{Context Tuning for In-Context Optimization}
\author{
  Jack Lu, Ryan Teehan, Zhenbang Yang, and Mengye Ren \\
  Agentic Learning AI Lab, New York University \\
  \texttt{\{yl11330, rst306, zy3101, mengye\}@nyu.edu}\\
  \url{https://agenticlearning.ai/context-tuning}
}
\date{July 3, 2026}
\maketitle

\begin{abstract}
\looseness=-10000
We introduce \ourmethod{}, a simple and effective method to significantly enhance few-shot adaptation of large language models (LLMs) without weight updates.
In-Context Learning (ICL) forms a memory representation of the demonstrations in a single forward pass but cannot refine it when insufficient.
Prompt-based methods offer lightweight adaptation by optimizing a trainable prompt or prefix but initialize it independently of the demonstrations.
In contrast, \ourmethod{} leverages the model's inherent ICL ability to initialize a trainable memory representation from demonstrations, then refines it through gradient-based optimization.
Extensive evaluations on benchmarks such as CrossFit, UnifiedQA, MMLU, BIG-Bench Hard, and ARC demonstrate that \ourmethod{} outperforms both ICL and traditional prompt-based adaptation methods while achieving competitive accuracy with Test-Time Training at significantly higher training efficiency.
\end{abstract}

\section{Introduction}\label{sec:intro}

\looseness=-10000
Large language models (LLMs) have demonstrated impressive capabilities across a wide range of natural language processing (NLP) tasks by leveraging knowledge acquired during large-scale pretraining~\citep{gpt3, llama3, mistral}. These models can adapt to new tasks using only a few input and output examples provided in context, a process known as In-Context Learning (ICL)~\citep{gpt3}. However, ICL often struggles with complex reasoning or domain shifts, as it relies solely on a single forward pass to form a memory representation of the examples. While methods like Test-Time Training (TTT)~\citep{ttt} have shown that effective adaptation is possible with limited data, they can still be computationally expensive. This highlights the need for more efficient and effective approaches to task adaptation in LLMs.

\looseness=-10000
Contrary to ICL's reliance on a forward pass, prompt-based adaptation methods like Prompt Tuning~\citep{prompttuning} and Prefix Tuning~\citep{prefixtuning} prepend a set of trainable vectors to each example input and optimize them via gradient descent. At a conceptual level, ICL harnesses the model's ability to extract task-relevant information from the context of few-shot examples, while prompt-based adaptation methods optimize randomly initialized vectors to guide the model's behavior in solving each example. Given these complementary approaches, it is natural to ask whether we can bridge them by directly optimizing the context representation formed from few-shot examples.

\begin{wrapfigure}{r}{0.5\textwidth}
  \centering
  \includegraphics[trim={0cm 0cm 0cm 0cm},clip,width=0.5\textwidth]{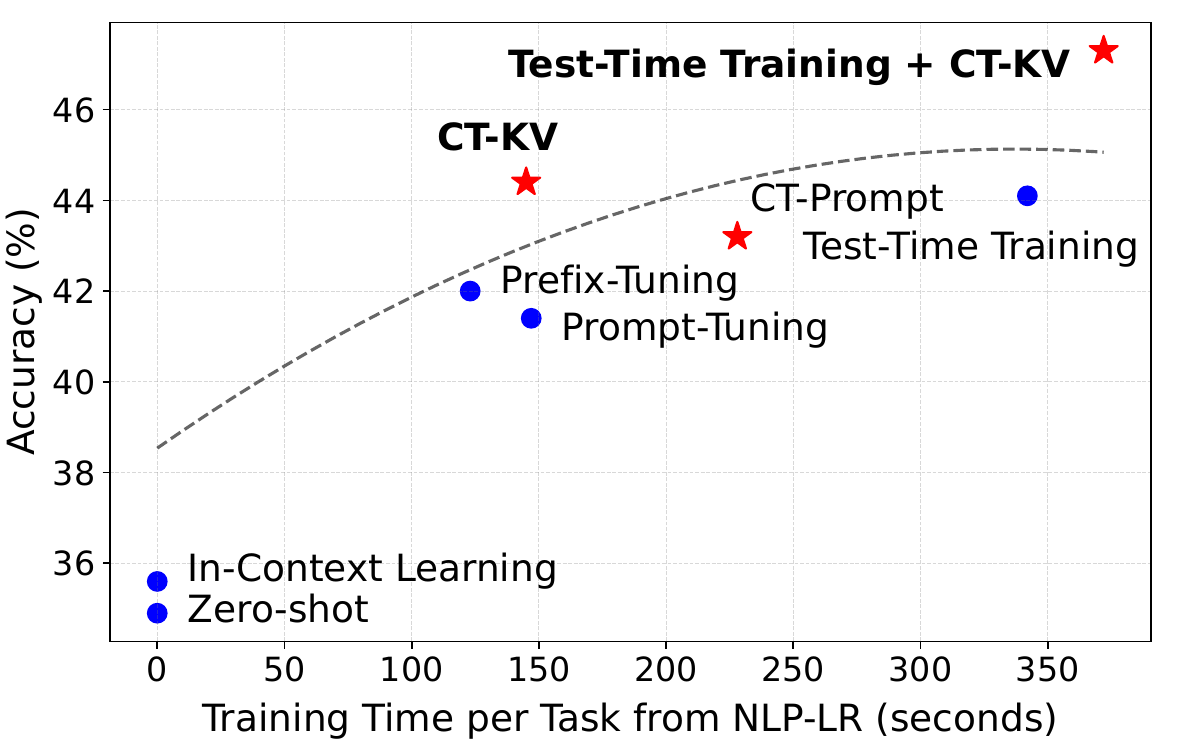}
  \caption{
    \looseness=-10000
    Comparison of training-free, prompt-based adaptation, and \ourframework{} methods on solving 26 NLP-LR tasks from Table~\ref{tab:main}. \textcolor{blue}{Circles} are baselines; \textcolor{red}{stars} are our methods; {\bf bolded} methods attain the best performance-efficiency tradeoff.
  }
  \label{fig:teaser}
\end{wrapfigure}

\looseness=-10000
In this work, we introduce \ourmethod{}, a simple and effective method for few-shot learning that initializes a trainable context representation from the few-shot examples of a novel task, then optimizes it to solve each example. We develop two variants: \ourmethodprompt{}, which applies Prompt Tuning to optimize a soft prompt initialized from few-shot examples, and \ourmethodkv{}, which optimizes the key-value (KV) cache derived from those same examples as the model's memory representation of those demonstrations. While \ourmethodprompt{} achieves strong performance, it suffers from a quadratic training-time cost in the number of examples. Similarly, the recently proposed Test-Time Training (TTT)~\citep{ttt} method, which fine-tunes model parameters with LoRA~\citep{lora} on permutations of few-shot examples, also incurs quadratic cost. In contrast, \ourmethodkv{} achieves linear training time complexity while also outperforming \ourmethodprompt{} and achieving competitive performance with TTT, due to the efficiency and per-layer conditioning of the KV cache. In addition, because \ourmethod{} tunes the context and TTT tunes the model, the two methods are complementary: applying \ourmethodkv{} to refine the model context after TTT's weight updates leads to additional performance gains. A high-level comparison in Figure~\ref{fig:teaser} illustrates \ourmethodkv{}'s high efficiency and accuracy, whether used alone or in combination with TTT.

\looseness=-10000
We situate two approaches for few-shot learning with in-context examples within a broader framework we term \ourframework{} (\ico{}): TTT optimizes the model itself, while \ourmethod{} optimizes the model's context. Under this framework, adaptation leverages the LLM's ICL ability and either updates its parameters or its context representation. We evaluate ICL, prompt-based adaptation methods, and \ico{} techniques across a wide range of natural language and symbolic reasoning benchmarks, including CrossFit~\citep{crossfit}, UnifiedQA~\citep{unifiedqa}, BIG-Bench Hard (BBH)~\citep{bbh, bbhsplit}, MMLU~\citep{mmlu}, and the Abstraction and Reasoning Corpus (ARC)~\citep{arcpaper}. \ourmethodkv{} significantly outperforms both ICL and prompt-based adaptation methods, while remaining competitive with the more computationally intensive TTT approach. Furthermore, we show that \ourmethodkv{} can serve as a post-hoc refinement step following TTT, leading to improved few-shot adaptation performance compared to either method used in isolation.

\section{Related Work}

\paragraph{Prompt-Based Adaptation.}
\looseness=-10000
Prompt-based adaptation steers pretrained language models to solve downstream tasks by learning task-specific inputs while keeping the model weights frozen. AutoPrompt~\citep{autoprompt} was an early method that constructed discrete prompts via gradient-based search. Prefix Tuning~\citep{prefixtuning} introduced trainable continuous vectors that serve as a prefix to the model's key-value cache at each layer, achieving strong performance on generation tasks with only a small number of trainable parameters. P-Tuning~\citep{ptuning} appended soft prompts to the input and used an LSTM-based prompt encoder to model dependencies between prompt tokens. Prompt Tuning~\citep{prompttuning} simplified the approach by learning soft prompts solely at the input layer and demonstrated that performance improves with model scale. P-Tuning v2~\citep{ptuningv2} extended prompt tuning to deeper layers for natural language understanding tasks. While these works typically initialize their learnable prompts using high-level task instructions, random tokens, or unrelated words, \ourmethod{} leverages the pretrained LLM's ability to extract meaningful task-specific information directly from in-context demonstration pairs. Finally,~\citet{ipt} proposed Instruction Prompt Tuning, in which expert-curated few-shot demonstrations are prepended to a learned soft prompt. In contrast, \ourmethod{} draws demonstration pairs directly from the dataset and uses them to initialize the prompt rather than prepending them as input.

\paragraph{In-Context Learning.}
\looseness=-10000
Popularized by~\citet{gpt3}, ICL has become a defining feature of large language models (LLMs), enabling them to perform novel tasks by conditioning on a few input-output demonstrations without any parameter updates. This behavior has been leveraged through various prompting strategies, such as Chain-of-Thought prompting to elicit reasoning~\citep{cot} and self-consistency decoding to reduce variance~\citep{selfconsistency}. Prior work has also explored selecting informative demonstrations~\citep{goodexamples, supportexample}, as well as meta-training over large sets of tasks to improve ICL generalization and inference-time efficiency~\citep{metaicl, metaicl2, metaprompt1, metaprompt2, metaprompt3, streamadapter}. From a theoretical perspective,~\citet{iclandsgd1} links ICL to implicit gradient descent, while~\citet{iclandsgd2} challenges this view;~\citet{iclashopfield} conceptualizes it as contextual retrieval within an associative memory framework; and~\citet{percyliangtoy} demonstrates that transformers trained from scratch can learn complex function classes in-context. While these findings highlight ICL's potential, recent studies by~\citet{rethinkingdemonstrations} and~\citet{demoshortcut} show that LLMs often only rely on superficial patterns in the demonstrations rather than learning the underlying task. Motivated by these limitations, \ourmethod{} uses gradient optimization to refine context representations initialized from demonstration pairs.

\paragraph{Inference-Time Optimization.}
\looseness=-10000
Our framework, \ourframework{}, contributes to a broader class of methods that adapt models or their internal representations at inference time. Originally applied to image classification~\citep{originalttt, tttmaskedauto}, test-time training has since shown strong results in language modeling~\citep{tttlanguage}, video generation~\citep{tttvideogen}, controllable language generation~\citep{icv}, and abstract reasoning~\citep{latentprogramsearch}. In diffusion models~\citep{diffusion, stablediffusion}, techniques such as classifier guidance and classifier-free guidance~\citep{classifierguidance, cfg} steer generation by optimizing intermediate outputs during sampling. These methods have enabled controllable text-to-image synthesis~\citep{glide}, adjustable aesthetic attributes~\citep{doodl}, and improved sample diversity~\citep{procreate}. More recently,~\citet{ttt} proposed test-time training of LoRA~\citep{lora} parameters for ICL using a leave-one-out strategy, achieving state-of-the-art performance on the Abstraction and Reasoning Corpus (ARC)~\citep{arcgithub, arcpaper}. In contrast, \ourmethod{} tunes a soft prompt or continuous prefix rather than updating model weights, and we evaluate it on a broader range of ICL tasks.

\section{Background}\label{sec:background}

\looseness=-10000
We introduce the mathematical formulation of ICL, Prompt Tuning, and Prefix Tuning. To set up the problem of single-task few-shot adaptation, we consider a language model $p_\phi$ with parameters $\phi$, $L$ layers, a demonstration set
\begin{equation*}
  \mathcal{D} = \{(x_i, y_i)\}_{i=1}^k,
\end{equation*}
\looseness=-10000
and the goal of solving a new query $x_q$ from the same task. We denote the concatenated context of all demonstration pairs as $\mathcal{C}=[x_1;y_1;\dots;x_k;y_k]$.

\paragraph{In-Context Learning.}
\looseness=-10000
ICL concatenates all $k$ demonstration pairs followed by the query $x_q$. The model then predicts $\hat y_q$ conditioned on this context:
\begin{equation*}
  \hat y_q = \arg\max_{y}\; p_{\phi}\bigl(y \bigm| [\,\mathcal{C}; x_q\,]\bigr).
\end{equation*}
\looseness=-10000
In ICL, there is no gradient-based optimization; instead, the model adapts by attending to the tokens of the demonstration pairs provided in context.

\paragraph{Prompt Tuning.}
\looseness=-10000
In Prompt Tuning, the model parameters $\phi$ remain fixed. Instead, $m$ trainable soft prompt tokens $P$ are prepended to each input and optimized via gradient descent:
\begin{equation}
  P^* = \arg\min_{P}
    \sum_{i=1}^{k}
    -\log p_{\phi}\bigl(y_i \bigm| [\,P; x_i\,]\bigr).
  \label{eqn:prompttuningoptimization}
\end{equation}
\looseness=-10000
After optimizing on the demonstration pairs, the optimized soft prompt $P^*$ can be used for inference:
\begin{equation*}
  \hat y_q = \arg\max_{y}\; p_{\phi}\bigl(y \bigm| [\,P^*; x_q\,]\bigr).
\end{equation*}

\paragraph{Prefix Tuning.}
\looseness=-10000
Prefix Tuning also keeps $\phi$ fixed but learns layer-wise prefixes of $m$ trainable vectors for the keys and values in each transformer layer:
\begin{equation*}
  \Theta = \{K_j, V_j\}_{j=1}^L.
\end{equation*}
\looseness=-10000
Each layer's attention uses these prefixes by prepending $K_j$ and $V_j$, each containing $m$ prefix vectors, to its keys and values. The prefixes are optimized to minimize
\begin{equation}
  \Theta^* = \arg\min_{\Theta}
    \sum_{i=1}^{k}
    -\log p_{\phi}\bigl(y_i \bigm| [\,\Theta; x_i\,]\bigr).
  \label{eqn:prefixtuningoptimization}
\end{equation}
\looseness=-10000
After obtaining $\Theta^*$, inference on the query $x_q$ proceeds analogously to Prompt Tuning.

\section{Context Tuning for In-Context Optimization}\label{sec:method}

\begin{figure}[t]
  \centering
  \includegraphics[trim={5cm 5.75cm 5cm 5cm},clip,width=1.0\textwidth]{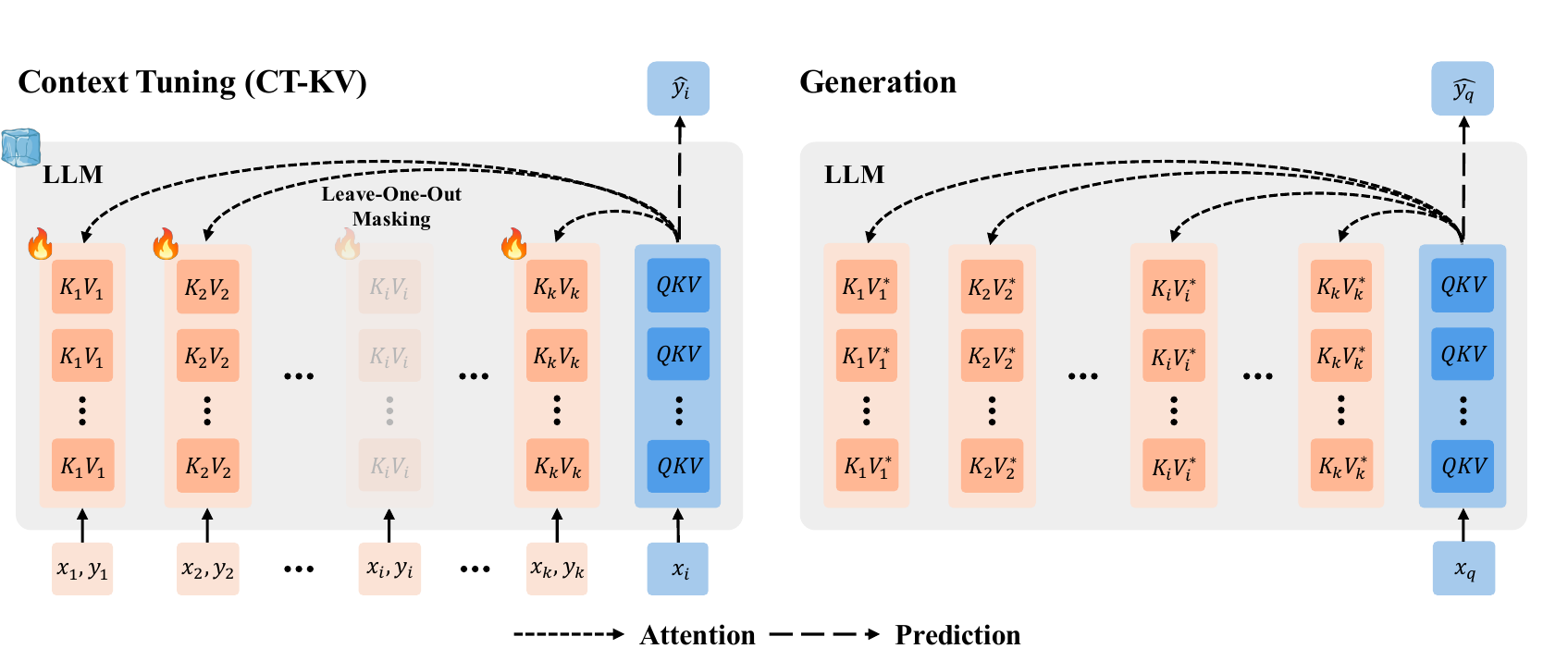}
  \caption{
    \looseness=-10000
    \ourmethodkv{}, the variant of \ourmethod{} that optimizes the key-value prefixes derived from in-context demonstration pairs. \ourmethodkv{} (left) first initializes a prefix $\{K_i, V_i\}_{i=1}^k$ from demonstration pairs $\{(x_i, y_i)\}_{i=1}^k$, then trains it to solve each pair. To prevent the model from simply retrieving the demonstration pair from the prefix, Leave-One-Out Masking prevents the model from attending to $K_i, V_i$ when solving pair $i$. At generation time (right), the model conditions on all optimized prefixes $\{K_i^*, V_i^*\}_{i=1}^k$ to solve query $x_q$.
  }
  \label{fig:main}
\end{figure}

\looseness=-10000
In this section, we introduce the mathematical formulation of \ourframework{} (\ico{}), a few-shot adaptation scheme that uses demonstrations in the context and performs gradient-based optimization on either the model parameters or a context representation derived from demonstrations. We then show that Test-Time Training (TTT)~\citep{ttt} is an instance of \ico{}. Finally, we present Context Tuning, formalizing its \ourmethodprompt{} and \ourmethodkv{} variants along with the two additional design choices that drive their strong performance. Figure~\ref{fig:main} illustrates the \ourmethodkv{} training and generation pipeline.

\subsection{In-Context Optimization}

\looseness=-10000
\ico{} combines ICL with gradient-based optimization to leverage the strengths of both. Formally, the objective of \ico{} is to minimize the loss
\begin{equation}
  \sum_{i=1}^{k}
    -\log p_\phi \left(
        y_i \bigm| [\,\theta_{\mathrm{context}}^{(i)}\, ; x_i\,]
    \right),
    \label{eqn:ico}
\end{equation}
\looseness=-10000
where $\theta_{\mathrm{context}}^{(i)}$ is a context representation derived from the set of demonstration pairs $\mathcal{D} = \{(x_i,y_i)\}_{i=1}^{k}$. One may notice that this objective resembles Equations~\ref{eqn:prompttuningoptimization} and~\ref{eqn:prefixtuningoptimization} because traditional prompt-based adaptation methods also prepend additional contexts to inputs during optimization. Still, these contexts are randomly initialized instead of utilizing the demonstration pairs $\mathcal{D}$. Therefore, Prompt Tuning and Prefix Tuning are not instances of \ico{} by definition. Furthermore, since ICL does not perform gradient-based optimization at all, it also does not fall under \ico{}.

\subsection{Test-Time Training as ICO}

\looseness=-10000
TTT~\citep{ttt} can be viewed as an instance of \ico{}. Specifically, TTT minimizes Equation~\ref{eqn:ico} by first initializing the model weights $\phi$ from a pretrained model, then updating them with LoRA layers for parameter efficiency. At each optimization iteration, TTT dynamically sets
\begin{equation}
\theta_{\mathrm{context}}^{(i)}=\mathcal{C}^{-i},
\end{equation}
where $\mathcal{C}^{-i}$ represents the concatenated tokens of a random permutation of demonstration pairs except for the $i$-th pair. Therefore, the optimization equation becomes:
\begin{equation}
  \phi^* = \arg\min_{\phi} \; \sum_{i=1}^{k}
    -\log p_\phi \left(
        y_i \bigm| [\,\mathcal{C}^{-i}\, ; x_i\,]
    \right).
\end{equation}
\looseness=-10000
To perform inference on the query input $x_q$, TTT uses the optimized model weights and the concatenation of all demonstration pairs as context:
\begin{equation}
    \hat{y}_q = \arg\max_{y}\,p_{\phi^{*}}\left(
        y \bigm| [\,\mathcal{C}\, ; x_q\,]
    \right).
\end{equation}

\subsection{Context Tuning}

\looseness=-10000
We design our \ourmethod{} approach to be an instantiation of the \ico{} framework. In contrast to TTT, \ourmethod{} freezes model parameters $\phi$ and instead directly optimizes the lightweight context representation $\theta_{\mathrm{context}}$ derived from the demonstration pairs.
\begin{itemize}
    \item \looseness=-10000 \ourmethodprompt{} initializes $\theta_{\mathrm{context}}=P_{\mathrm{CT}}$ as the model's prompt embeddings on $\mathcal{C}$, the concatenation of demonstration pairs.
    \item \looseness=-10000 \ourmethodkv{} initializes $\theta_{\mathrm{context}}=\Theta_{\mathrm{CT}}$ as a key-value prefix $\Theta_{\mathrm{CT}}=\{K_j, V_j\}_{j=1}^{L}$ obtained from the model's layer-wise activations on $\mathcal{C}$.
\end{itemize}
\looseness=-10000
Furthermore, we introduce two design choices for both \ourmethodprompt{} and \ourmethodkv{}. We study the performance impact of each in Section~\ref{sec:ablation}, demonstrating that both are crucial for achieving strong empirical gains.

\paragraph{Leave-One-Out Masking.}
\looseness=-10000
To prevent the model from simply retrieving the answer $y_i$ of the $i$-th demonstration pair embedded in $\theta_{\mathrm{context}}$ when predicting the output for $x_i$, we construct
\begin{equation}
  \theta_{\mathrm{context}}^{(i)} =
  \begin{cases}
      P_{\mathrm{CT}}^{-i} & \text{for \ourmethodprompt{}},\\
      \Theta_{\mathrm{CT}}^{-i} & \text{for \ourmethodkv{}},
  \end{cases}
\end{equation}
\looseness=-10000
and use it instead of $\theta_{\mathrm{context}}$ in optimization. When conditioning on $P_{\mathrm{CT}}^{-i}$ or $\Theta_{\mathrm{CT}}^{-i}$, the trainable soft prompt tokens in \ourmethodprompt{} or prefix tokens in \ourmethodkv{} corresponding to the in-context demonstration pair $(x_i,y_i)$ are masked out from the attention view of the model by setting the corresponding positions of the attention mask to $0$. In contrast to TTT's leave-one-out technique, which omits one demonstration pair in the context to update the model weights, our Leave-One-Out Masking operates on the derived context vectors with the model parameters frozen, ensuring that the optimization refines the context representation itself rather than relying on weight updates.

\paragraph{Token Dropout.}
\looseness=-10000
Since \ourmethod{} generally introduces a larger number of prompt or prefix tokens than traditional prompt-based adaptation techniques, we regularize training by randomly dropping tokens in $\theta_{\mathrm{context}}^{(i)}$ with a fixed probability, denoted as $\mathrm{TokenDrop}$. During optimization, the loss is computed in expectation over these stochastic dropout masks, encouraging the learned context to avoid overfitting to any single token.

\noindent
\looseness=-10000
Altogether, we arrive at the optimization equations for \ourmethodprompt{} and \ourmethodkv{}:
\begin{align}
    \text{\ourmethodprompt{}:}\quad& P_{\mathrm{CT}}^*
      = \arg\min_{P_{\mathrm{CT}}}
         \sum_{i=1}^{k}
         -\log p_\phi\left(y_i \bigm| [\,\mathrm{TokenDrop}\left(P_{\mathrm{CT}}^{-i}\right)\,;x_i]\right), \\
    \text{\ourmethodkv{}:}\quad& \Theta_{\mathrm{CT}}^*
      = \arg\min_{\Theta_{\mathrm{CT}}}
         \sum_{i=1}^{k}
         -\log p_\phi\left(y_i \bigm| [\,\mathrm{TokenDrop}\left(\Theta_{\mathrm{CT}}^{-i}\right)\,;x_i]\right).
\end{align}

\noindent
\looseness=-10000
To perform inference on the query $x_q$, \ourmethodprompt{} and \ourmethodkv{} use their respective optimized contexts:
\begin{align}
  \text{\ourmethodprompt{}:}\quad&
  \hat{y}_q = \arg\max_{y}\,p_{\phi}\left(
        y \bigm| [\,P_{\mathrm{CT}}^*\, ; x_q\,]
  \right), \\
  \text{\ourmethodkv{}:}\quad&
  \hat{y}_q = \arg\max_{y}\,p_{\phi}\left(
        y \bigm| [\,\Theta_{\mathrm{CT}}^*\, ; x_q\,]
  \right).
\end{align}

\looseness=-10000
In practice, \ourmethodprompt{} requires recomputing layer-wise keys and values corresponding to $P_{\mathrm{CT}}$, while \ourmethodkv{} does not. In Appendix~\ref{sec:timecomplexity}, we formally prove that for each optimization step, \ourmethodkv{} has lower time complexity than both TTT and \ourmethodprompt{} with respect to the number of demonstration pairs. Finally, we introduce \tttct{}, which first performs TTT to update model weights $\phi$, then applies \ourmethodkv{} to refine the model's demonstration context for improved performance.

\section{Experiments}\label{sec:experiment}

\subsection{Datasets}

\begin{figure}[!t]
  \centering
  \includegraphics[trim={4.25cm 2.5cm 4.2cm 2.3cm},clip,width=\textwidth]{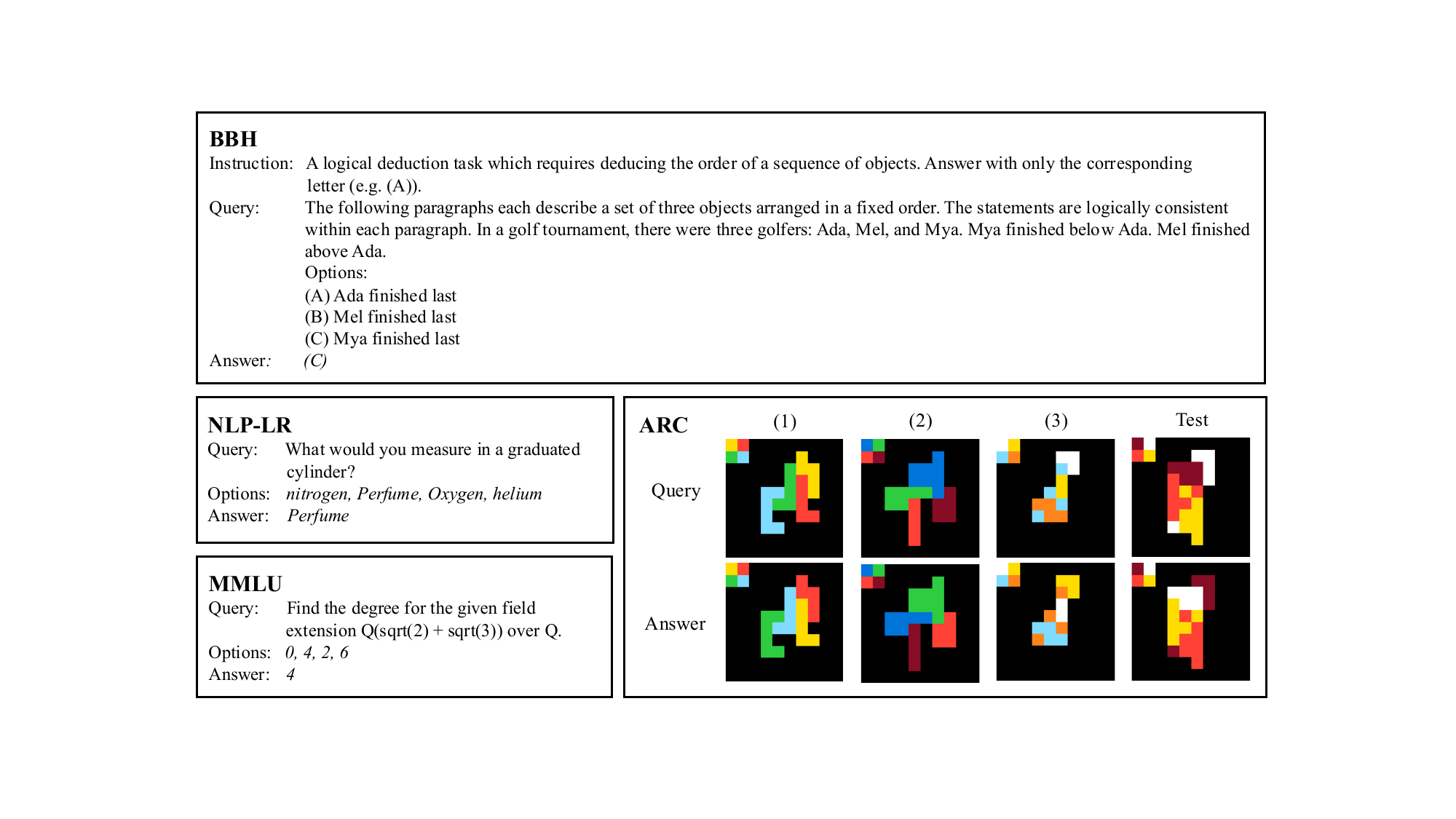}
  \caption{
    \looseness=-10000
    One test pair from BBH, NLP-LR, and MMLU each, and three demonstration pairs followed by a test pair from ARC. BBH contains instructions that we prepend to model inputs. NLP-LR and MMLU contain multiple-choice options for the model to select. To avoid clutter, we show demonstration pairs from BBH, NLP-LR, and MMLU in Appendix~\ref{appendix:qualitative}.
  }
  \label{fig:dataset}
\end{figure}

\looseness=-10000
We evaluate on a diverse set of challenging datasets for pretrained LLMs. We show a representative task example for each dataset in Figure~\ref{fig:dataset}.
\begin{itemize}
    \item \looseness=-10000 {\bf NLP-LR} is the low-resource dataset split introduced by~\citet{metaicl}, encompassing 26 NLP tasks from CrossFit~\citep{crossfit} and UnifiedQA~\citep{unifiedqa}, such as sentiment analysis and paraphrasing. Following~\citet{metaicl}, we sample $k=16$ demonstration pairs per task and evaluate task instances as multiple-choice problems.
    \item \looseness=-10000 {\bf Massive Multitask Language Understanding (MMLU)} is a diverse benchmark consisting of 57 subject-specific tasks, including mathematics, history, law, and various other domains~\citep{mmlu}. We sample $k=16$ demonstration pairs per task and evaluate task instances as multiple-choice problems.
    \item \looseness=-10000 {\bf BIG-Bench Hard (BBH)} is a curated subset of BIG-Bench, consisting of 23 tasks that challenge pretrained LLMs with questions involving algorithmic puzzles, symbolic manipulation, and other complex reasoning domains~\citep{bbh, bbhsplit}. Following~\citet{ttt}, we sample $k=10$ demonstration pairs per task and prepend task instructions to model inputs. We evaluate these tasks as question-answering problems.
    \item \looseness=-10000 {\bf Abstraction and Reasoning Corpus (ARC)} is a challenging symbolic reasoning benchmark with 400 evaluation tasks, each defined by a few grid transformation pairs and one or more query input grids~\citep{arcpaper}. Since the average number of available demonstration pairs is fewer than 4, we use all of them in context. Tasks are evaluated as question-answering problems.
\end{itemize}

\looseness=-10000
Each dataset is formatted either as a multiple-choice task or a question-answering task. For multiple-choice problems, where the LLM must select an output from a predefined set of answers, we follow~\citet{metaicl} and choose the option with the lowest loss. For question-answering tasks, the LLM has to generate an answer that matches the ground-truth output.

\subsection{Models}

\looseness=-10000
Following~\citet{metaicl} and~\citet{ttt}, we use GPT-2~\citep{gpt2} and Llama 3-8B~\citep{llama3} for NLP-LR and BBH, respectively. To demonstrate that \ourmethod{} performs well across different model sizes, we select Llama 3.2-3B for MMLU. Due to computational constraints, we use Llama 3.2-1B for ARC, which requires handling long input sequences. Since the pretrained Llama 3.2-1B model cannot solve any of the 400 ARC evaluation tasks, we follow~\citet{ttt} and~\citet{architect} by fine-tuning it on the ARC training split, which contains 400 tasks that do not overlap with the evaluation split. Prior ARC-focused work includes systems targeting high ARC competition scores~\citep{arcreport}, such as~\citet{architect}, and studies of test-time adaptation on ARC~\citep{ttt}. Our goal is to develop a method applicable across general few-shot problems, so we do not perform augmentation or voting for ARC. All pretrained model checkpoints were obtained from Hugging Face.

\subsection{Experiment Setup}\label{sec:experimentsetup}

\looseness=-10000
On top of zero-shot inference and ICL, we compare a variety of few-shot learning techniques to conduct a broad investigation of prompt-based adaptation strategies and methods under the \ourframework{} framework: Prompt Tuning, Prefix Tuning, TTT, \ourmethodprompt{}, \ourmethodkv{}, and \tttct{}. We use greedy decoding for all question-answering tasks. All experiments in Table~\ref{tab:main} are run over 5 different sets of randomly selected demonstration pairs, except for ARC, which has a fixed set of demonstration pairs for each task.

\looseness=-10000
For Prompt Tuning and Prefix Tuning, we either set the number of trainable tokens $m$ to 32, or match it to the number of tokens that are in the demonstration pairs used by \ourmethod{}. Trainable soft prompts and prefixes are initialized using sampled token embeddings from the model, which we find yields the best baseline performance.

\looseness=-10000
For ARC, we fine-tune our Llama 3.2-1B checkpoint following the setup of~\citet{architect}, using 2 A100 GPUs for 24 epochs with a learning rate of $2\times 10^{-4}$, a cosine learning rate scheduler, 1 warm-up epoch, and a global batch size of 32 (after gradient accumulation). All other experiments are conducted on a single A100 GPU, except NLP-LR, which is run on an RTX 8000. For \ourmethodprompt{} and \ourmethodkv{}, we apply Leave-One-Out Masking from Section~\ref{sec:method} across all datasets, except ARC, where performance improves without it. We elaborate on this decision in Section~\ref{sec:ablation}.

\looseness=-10000
For completeness, we also compare alternative setups for both Prompt Tuning and Prefix Tuning. In Appendix~\ref{appendix:moreinit}, we report results using uniformly initialized trainable parameters for both methods. We also include results for Prefix Tuning with an MLP parameterization, along with details of our hyperparameter search to support reproducibility. Overall, our evaluation spans a wide range of challenging tasks, model sizes from 1B to 8B parameters, varied numbers of demonstration pairs per task ($k = 2$ to $k = 16$), and benchmarks with and without task instructions (e.g., BBH includes instructions, while the others do not).

\label{sec:results}

\subsection{Comparing Context Tuning to Baselines}

\begin{table}[t]
  \centering
  \setlength{\tabcolsep}{2pt}
  \small
  \caption{
    \looseness=-10000
    Few-shot learning performance on NLP-LR, MMLU, BBH, and ARC benchmarks. For each benchmark, we report accuracy (\%) and training time per task (seconds). We show the means and \textcolor{gray}{standard deviations} of accuracies over 5 seeds with different sets of demonstration pairs per task (except ARC because it has fixed demonstration pairs). The best accuracy is \best{bolded} and the second best is \secondbest{underlined} for each benchmark.
  }
  \begin{tabular}{@{}L{0.305\linewidth}|C{0.105\linewidth}C{0.05\linewidth}|C{0.105\linewidth}C{0.05\linewidth}|C{0.105\linewidth}C{0.05\linewidth}|C{0.105\linewidth}C{0.05\linewidth}@{}}
    \toprule
    \textbf{Method}
      & \multicolumn{2}{c|}{\textbf{NLP-LR}}
      & \multicolumn{2}{c|}{\textbf{MMLU}}
      & \multicolumn{2}{c|}{\textbf{BBH}}
      & \multicolumn{2}{c}{\textbf{ARC}} \\
    \cmidrule(lr){2-3} \cmidrule(lr){4-5} \cmidrule(lr){6-7} \cmidrule(lr){8-9}
      & \multicolumn{1}{c}{\scriptsize\textbf{Acc.\,(\%)}} & \multicolumn{1}{c|}{\scriptsize\textbf{T (s)}}
      & \multicolumn{1}{c}{\scriptsize\textbf{Acc.\,(\%)}} & \multicolumn{1}{c|}{\scriptsize\textbf{T (s)}}
      & \multicolumn{1}{c}{\scriptsize\textbf{Acc.\,(\%)}} & \multicolumn{1}{c|}{\scriptsize\textbf{T (s)}}
      & \multicolumn{1}{c}{\scriptsize\textbf{Acc.\,(\%)}} & \multicolumn{1}{c}{\scriptsize\textbf{T (s)}} \\
    \midrule
    \multicolumn{9}{l}{\textbf{Baselines}} \\
    \rowwhite Zero-Shot
      & $34.9{\scriptstyle\,\pm\,\textcolor{gray}{0.62}}$ & $0$
      & $35.8{\scriptstyle\,\pm\,\textcolor{gray}{0.71}}$ & $0$
      & $40.9{\scriptstyle\,\pm\,\textcolor{gray}{0.43}}$ & $0$
      & $1.0$ & $0$ \\
    \rowwhite ICL
      & $35.6{\scriptstyle\,\pm\,\textcolor{gray}{0.65}}$ & $0$
      & $41.2{\scriptstyle\,\pm\,\textcolor{gray}{0.57}}$ & $0$
      & $50.4{\scriptstyle\,\pm\,\textcolor{gray}{0.78}}$ & $0$
      & $13.3$ & $0$ \\
    \rowwhite LoRA
      & $42.8{\scriptstyle\,\pm\,\textcolor{gray}{0.88}}$ & $156$
      & $40.1{\scriptstyle\,\pm\,\textcolor{gray}{0.93}}$ & $16$
      & $51.7{\scriptstyle\,\pm\,\textcolor{gray}{0.66}}$ & $9$
      & $13.5$ & $14$ \\
    \rowwhite Rank-Stabilized LoRA
      & $41.7{\scriptstyle\,\pm\,\textcolor{gray}{0.84}}$ & $143$
      & $38.8{\scriptstyle\,\pm\,\textcolor{gray}{0.87}}$ & $17$
      & $46.7{\scriptstyle\,\pm\,\textcolor{gray}{1.24}}$ & $8$
      & $12.8$ & $15$ \\
    \rowwhite DoRA
      & $42.9{\scriptstyle\,\pm\,\textcolor{gray}{0.87}}$ & $161$
      & $40.3{\scriptstyle\,\pm\,\textcolor{gray}{0.90}}$ & $16$
      & $52.6{\scriptstyle\,\pm\,\textcolor{gray}{0.86}}$ & $9$
      & $13.0$ & $15$ \\
    \rowwhite Prompt Tuning (m\,=\,32)
      & $41.4{\scriptstyle\,\pm\,\textcolor{gray}{1.02}}$ & $147$
      & $39.2{\scriptstyle\,\pm\,\textcolor{gray}{1.04}}$ & $15$
      & $50.8{\scriptstyle\,\pm\,\textcolor{gray}{1.59}}$ & $7$
      & $12.0$ & $13$
      \\
    \rowwhite Prompt Tuning (m\,=\,\# demo)
      & $38.8{\scriptstyle\,\pm\,\textcolor{gray}{1.23}}$ & $231$
      & $37.3{\scriptstyle\,\pm\,\textcolor{gray}{1.23}}$ & $29$
      & $47.5{\scriptstyle\,\pm\,\textcolor{gray}{1.84}}$ & $16$
      & $14.5$ & $49$ \\
    \rowwhite Prefix Tuning (m\,=\,32)
      & $42.0{\scriptstyle\,\pm\,\textcolor{gray}{0.85}}$ & $123$
      & $39.9{\scriptstyle\,\pm\,\textcolor{gray}{0.94}}$ & $5$
      & $52.7{\scriptstyle\,\pm\,\textcolor{gray}{1.12}}$ & $7$
      & $9.3$ & $14$ \\
    \rowwhite Prefix Tuning (m\,=\,\# demo)
      & $41.1{\scriptstyle\,\pm\,\textcolor{gray}{0.89}}$ & $144$
      & $38.8{\scriptstyle\,\pm\,\textcolor{gray}{0.81}}$ & $8$
      & $52.8{\scriptstyle\,\pm\,\textcolor{gray}{1.15}}$ & $9$
      & $20.5$ & $24$ \\
    \rowwhite TTT
      & $44.1{\scriptstyle\,\pm\,\textcolor{gray}{0.65}}$ & $342$
      & $43.6{\scriptstyle\,\pm\,\textcolor{gray}{0.55}}$ & $30$
      & $57.8{\scriptstyle\,\pm\,\textcolor{gray}{1.13}}$ & $14$
      & $\secondbest{23.8}$ & $56$ \\
    \midrule
    \multicolumn{9}{l}{\textbf{Our Methods}} \\
    \rowwhite \ourmethodprompt{}
      & $43.2{\scriptstyle\,\pm\,\textcolor{gray}{0.61}}$ & $228$
      & $43.6{\scriptstyle\,\pm\,\textcolor{gray}{0.67}}$ & $33$
      & $56.3{\scriptstyle\,\pm\,\textcolor{gray}{0.98}}$ & $14$
      & $22.5$ & $52$ \\
    \rowwhite \ourmethodkv{}
      & $\secondbest{44.2}{\scriptstyle\,\pm\,\textcolor{gray}{0.55}}$ & $145$
      & $\secondbest{43.7}{\scriptstyle\,\pm\,\textcolor{gray}{0.54}}$ & $9$
      & $\secondbest{57.9}{\scriptstyle\,\pm\,\textcolor{gray}{0.78}}$ & $7$
      & $\secondbest{23.8}$ & $26$ \\
    \rowblue \tttct{}
      & $\best{47.6}{\scriptstyle\,\pm\,\textcolor{gray}{0.53}}$ & $372$
      & $\best{44.1}{\scriptstyle\,\pm\,\textcolor{gray}{0.38}}$ & $34$
      & $\best{58.2}{\scriptstyle\,\pm\,\textcolor{gray}{0.73}}$ & $17$
      & $\best{25.8}$ & $63$ \\
    \bottomrule
  \end{tabular}
  \label{tab:main}
\end{table}

\looseness=-10000
Table~\ref{tab:main} reports the performance and training time per task for our baselines and methods across the four benchmarks. To fairly compare Prompt Tuning and Prefix Tuning with \ourmethod{}, ``Prompt Tuning (m\,=\,\# demo)'' and ``Prefix Tuning (m\,=\,\# demo)'' are configured to match the number of trainable parameters in \ourmethodprompt{} and \ourmethodkv{}, respectively, by setting $m$ to the number of demonstration-pair tokens. We report each method's number of trainable parameters in Appendix~\ref{appendix:numparams}.

\paragraph{Context Tuning outperforms Prompt Tuning, Prefix Tuning, and LoRA variants.}
\looseness=-10000
\ourmethodprompt{} outperforms Prompt Tuning (m\,=\,32), and \ourmethodkv{} outperforms Prefix Tuning (m\,=\,32), both by a wide margin across all benchmarks. Moreover, increasing $m$ to match the number of demonstration tokens does not yield consistent improvements in Prompt Tuning or Prefix Tuning. Despite tuning the same number of parameters, these variants still underperform compared to \ourmethodprompt{} and \ourmethodkv{}. In addition, \ourmethodkv{} outperforms LoRA, rank-stabilized LoRA~\citep{rslora}, and DoRA~\citep{dora}. This highlights the effectiveness of leveraging the model's ICL capabilities by initializing the prompt or prefix with demonstration tokens.

\paragraph{CT-KV is more efficient than CT-Prompt.}
\looseness=-10000
\ourmethodkv{} exhibits significantly lower training time per task compared to \ourmethodprompt{}. This observation aligns with the time complexity discussion in Appendix~\ref{sec:timecomplexity}: \ourmethodprompt{} incurs quadratic scaling in training time with the number of demonstration pairs, while \ourmethodkv{} scales linearly. In addition to being faster, \ourmethodkv{} also outperforms \ourmethodprompt{} in accuracy by conditioning each transformer layer's activations with layer-specific key and value vectors, rather than relying solely on input-level soft prompts.

\paragraph{CT-KV offers an efficient alternative to TTT, and the two are complementary.}
\looseness=-10000
\ourmethodkv{} achieves performance comparable to TTT across NLP-LR, MMLU, and BBH, and solves the same number of ARC tasks. However, it requires at most half the training time per task compared to TTT on all benchmarks. This demonstrates that while the two methods converge to similar performance levels, \ourmethodkv{} is more efficient due to its linear time complexity in the number of demonstration pairs, in contrast to TTT's quadratic time complexity. Furthermore, \ourmethodkv{} can be applied as a refinement step after TTT updates the model weights, leading to higher performance on all benchmarks with minimal additional training time. This suggests that context and model-based adaptation methods within the \ourframework{} framework are complementary and can be effectively combined for few-shot learning.

\paragraph{Initialization from demonstration pairs lowers standard deviation in performance.}
\looseness=-10000
Initializing the trainable prompt or prefix from demonstration pairs, rather than from random tokens, reduces variance across random seeds in both \ourmethodprompt{} and \ourmethodkv{}. This leads to more stable performance compared to Prompt Tuning and Prefix Tuning.

\paragraph{CT-KV outperforms MetaICL on NLP-LR.}
\looseness=-10000
\ourmethodkv{} achieves 44.2\% accuracy on NLP-LR, surpassing the reported 43.3\% accuracy of MetaICL on NLP-LR~\citep{metaicl} when controlled for the same training and evaluation samples. This demonstrates that inference-time, single-task optimization with \ourmethodkv{} can rival the performance of approaches that fine-tune model weights across many tasks.
\begin{figure}[t]
  \centering
  \includegraphics[trim={0.6cm 5.5cm 0.6cm 5.5cm},clip,width=1.0\textwidth]{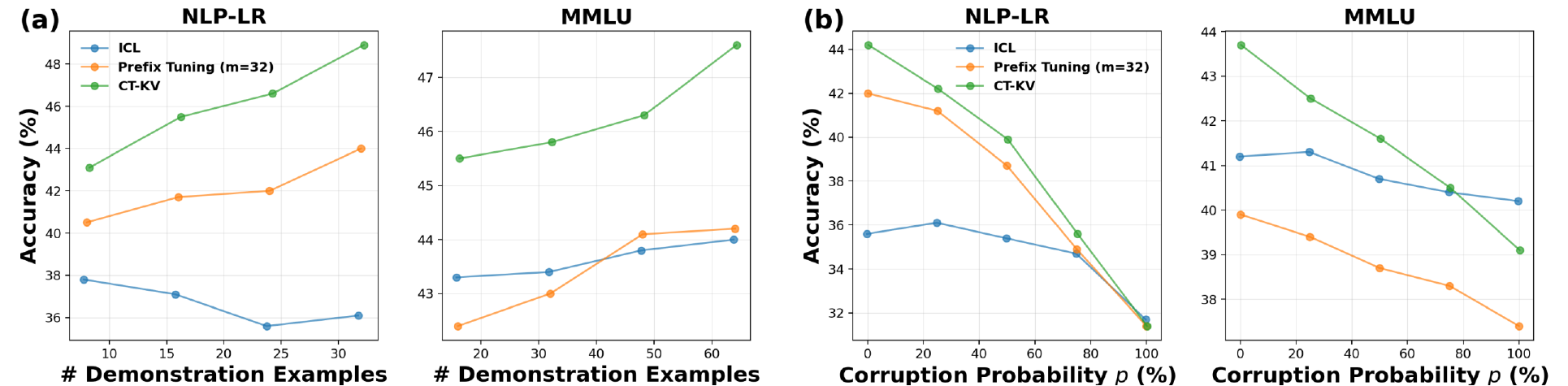}
  \caption{
    \looseness=-10000
    Performance of ICL, Prefix Tuning, and \ourmethodkv{} as (a) the number of demonstration pairs and (b) the demonstration-label corruption probability vary on NLP-LR and MMLU.
  }
  \label{fig:rebuttal}
\end{figure}

\subsection{Robustness to Demonstration Count and Quality}\label{sec:robustness}

\looseness=-10000
We now ablate the number of demonstration pairs ($k$) to assess \ourmethod{}'s robustness to it. We compare \ourmethodkv{} to ICL and Prefix Tuning on NLP-LR and MMLU, varying $k=4, 8, 12, 16$ for NLP-LR and $k=16, 32, 48, 64$ for MMLU. Figure~\ref{fig:rebuttal}(a) shows that across both benchmarks, \ourmethodkv{} consistently outperforms ICL and Prefix Tuning across $k$ values and attains strong performance gains as $k$ increases. Notably, despite ICL's limited scalability with $k$ on NLP-LR, \ourmethodkv{}'s accuracy improves more as $k$ increases than that of ICL or Prefix Tuning. We further evaluate \ourmethodkv{} on BBH in the many-shot regime~\citep{manyshoticl} at $k=25, 50, 100$ in Appendix~\ref{appendix:manyshot}, where its accuracy gap over ICL grows as $k$ scales.

\looseness=-10000
Next, we investigate \ourmethodkv{}'s robustness to low-quality demonstration examples by injecting label noise. For NLP-LR and MMLU, we corrupt demonstration examples by replacing the multiple-choice option with an incorrect option with a corruption probability $p$. Figure~\ref{fig:rebuttal}(b) shows that \ourmethodkv{} achieves the best performance across both benchmarks for corruption rates up to $75\%$, demonstrating strong robustness to noisy demonstration examples.

\FloatBarrier

\subsection{Scaling Up the Pretrained Models}\label{sec:scalingup}

\looseness=-10000
To broaden our model selections, we additionally evaluate \ourmethodkv{}, ICL, and Prefix Tuning on BIG-Bench Hard (BBH) with large, recently released pretrained models: Mistral-NeMo-12B-Instruct~\citep{mistralnemo}, DeepSeek-R1-Distill-Qwen-14B and its 32B variant~\citep{deepseek}, and Qwen3-14B and its 32B variant~\citep{qwen3}. Table~\ref{tab:scaling} shows that \ourmethodkv{} significantly outperforms both ICL and Prefix Tuning, demonstrating its effectiveness across modern LLMs of varying sizes and architectures. Notably, \ourmethodkv{}'s absolute accuracy gain over ICL reaches $6.1\%$ (Mistral-12B), with consistent improvements across all five tested models.

\begin{table}[h]
    \centering
    \setlength{\tabcolsep}{4pt}
    \small
    \caption{
        \looseness=-10000
        Performance across large-scale pretrained models.
    }
    \begin{tabular}{lccccc}
        \toprule
        \textbf{Method} & \textbf{Mistral-12B} & \textbf{DeepSeek-14B} & \textbf{DeepSeek-32B} & \textbf{Qwen3-14B} & \textbf{Qwen3-32B} \\
        \midrule
        \rowwhite ICL & $51.3$ & $55.2$ & $63.5$ & $60.8$ & $63.1$ \\
        \rowwhite Prefix Tuning (m\,=\,32) & $37.0$ & $49.8$ & $53.0$ & $49.3$ & $56.4$ \\
        \rowblue \ourmethodkv{} & $\best{57.4}$ & $\best{58.4}$ & $\best{66.7}$ & $\best{64.2}$ & $\best{69.1}$ \\
        \bottomrule
    \end{tabular}
    \label{tab:scaling}
\end{table}

\FloatBarrier

\subsection{Ablating Our Design Choices}\label{sec:ablation}

\looseness=-10000
We ablate two \ourmethodkv{} design choices: Leave-One-Out Masking and Token Dropout. Table~\ref{tab:ablation} shows that across all benchmarks, \ourmethodkv{} without Token Dropout performs marginally worse than \ourmethodkv{} with both components. This suggests that when tuning more parameters than traditional Prefix Tuning, applying dropout along the token dimension of $\Theta_{\mathrm{CT}}$ serves as an effective regularization technique for improving generalization. For NLP-LR, BBH, and MMLU, \ourmethodkv{} performs significantly worse when Leave-One-Out Masking is not applied. This indicates that during training, it is crucial to mask out the portion of $\theta_{\mathrm{context}}$ corresponding to the demonstration pair being solved, as it prevents the model from bypassing the task by retrieving the target output directly from the prefix initialization. However, on ARC, the model performs better without Leave-One-Out Masking. We hypothesize this is because ARC evaluation tasks typically include very few demonstration pairs (fewer than 4), so masking out even one pair during training can meaningfully reduce the effectiveness of the prompt or prefix in ICL. We also observe that when neither Leave-One-Out Masking nor Token Dropout is applied, \ourmethodkv{} performs worse than ICL on MMLU and only marginally better on BBH, highlighting that these two design choices are essential to its overall performance. These design choices also act as regularizers that keep \ourmethodkv{} stable under extended training, whereas Prefix Tuning overfits (Appendix~\ref{appendix:overfit}).

\begin{table}[ht]
  \centering
  \setlength{\tabcolsep}{4pt}
  \small
  \caption{
    \looseness=-10000
    Ablation study on the effects of Leave-One-Out Masking and Token Dropout in \ourmethodkv{}. Means and standard deviations are computed over 5 seeds, except for ARC, which has fixed demonstration pairs.
  }
  \begin{tabular}{l|rrrr}
    \toprule
    \textbf{Method}
      & \multicolumn{1}{c}{\textbf{NLP-LR}}
      & \multicolumn{1}{c}{\textbf{MMLU}}
      & \multicolumn{1}{c}{\textbf{BBH}}
      & \multicolumn{1}{c}{\textbf{ARC}} \\
    \midrule
    Neither
      & $41.0{\scriptstyle\,\pm\,\textcolor{gray}{0.75}}$
      & $40.2{\scriptstyle\,\pm\,\textcolor{gray}{0.73}}$
      & $51.4{\scriptstyle\,\pm\,\textcolor{gray}{0.76}}$
      & $21.0$ \\
    No Leave-One-Out Masking
      & $42.6{\scriptstyle\,\pm\,\textcolor{gray}{0.45}}$
      & $41.5{\scriptstyle\,\pm\,\textcolor{gray}{0.65}}$
      & $54.4{\scriptstyle\,\pm\,\textcolor{gray}{0.88}}$
      & $\best{23.8}$ \\
    No Token Dropout
      & $43.9{\scriptstyle\,\pm\,\textcolor{gray}{0.62}}$
      & $42.7{\scriptstyle\,\pm\,\textcolor{gray}{0.62}}$
      & $55.3{\scriptstyle\,\pm\,\textcolor{gray}{0.72}}$
      & $21.0$ \\
    Both
      & $\best{44.2}{\scriptstyle\,\pm\,\textcolor{gray}{0.55}}$
      & $\best{43.7}{\scriptstyle\,\pm\,\textcolor{gray}{0.54}}$
      & $\best{57.9}{\scriptstyle\,\pm\,\textcolor{gray}{0.78}}$
      & $22.5$ \\
    \bottomrule
  \end{tabular}
  \label{tab:ablation}
\end{table}

\FloatBarrier

\subsection{Qualitative Results}\label{sec:qualitative}

\looseness=-10000
We select an ARC task to illustrate how autoregressively generated answers improve across iterations. Figure~\ref{fig:qualitative} shows that at iteration $0$, which is equivalent to ICL, the model already identifies that red should be used to complete the cross shapes, but does not understand that it should avoid overwriting black squares. By iteration $200$, the prediction becomes more consistent with the demonstration pairs and solves the task. This example illustrates how optimizing the KV cache can refine the memory representation induced by the demonstrations, rather than relying on a single forward pass through the context. Appendix~\ref{appendix:qualitative_iterations} provides a complementary ARC example where \ourmethodkv{} corrects an initially wrong color mapping.

\begin{center}
  \captionsetup{type=figure,hypcap=false,skip=4pt}
  \includegraphics[trim={11.57cm 1.71cm 12.03cm 18.25cm},clip,width=0.9\textwidth]{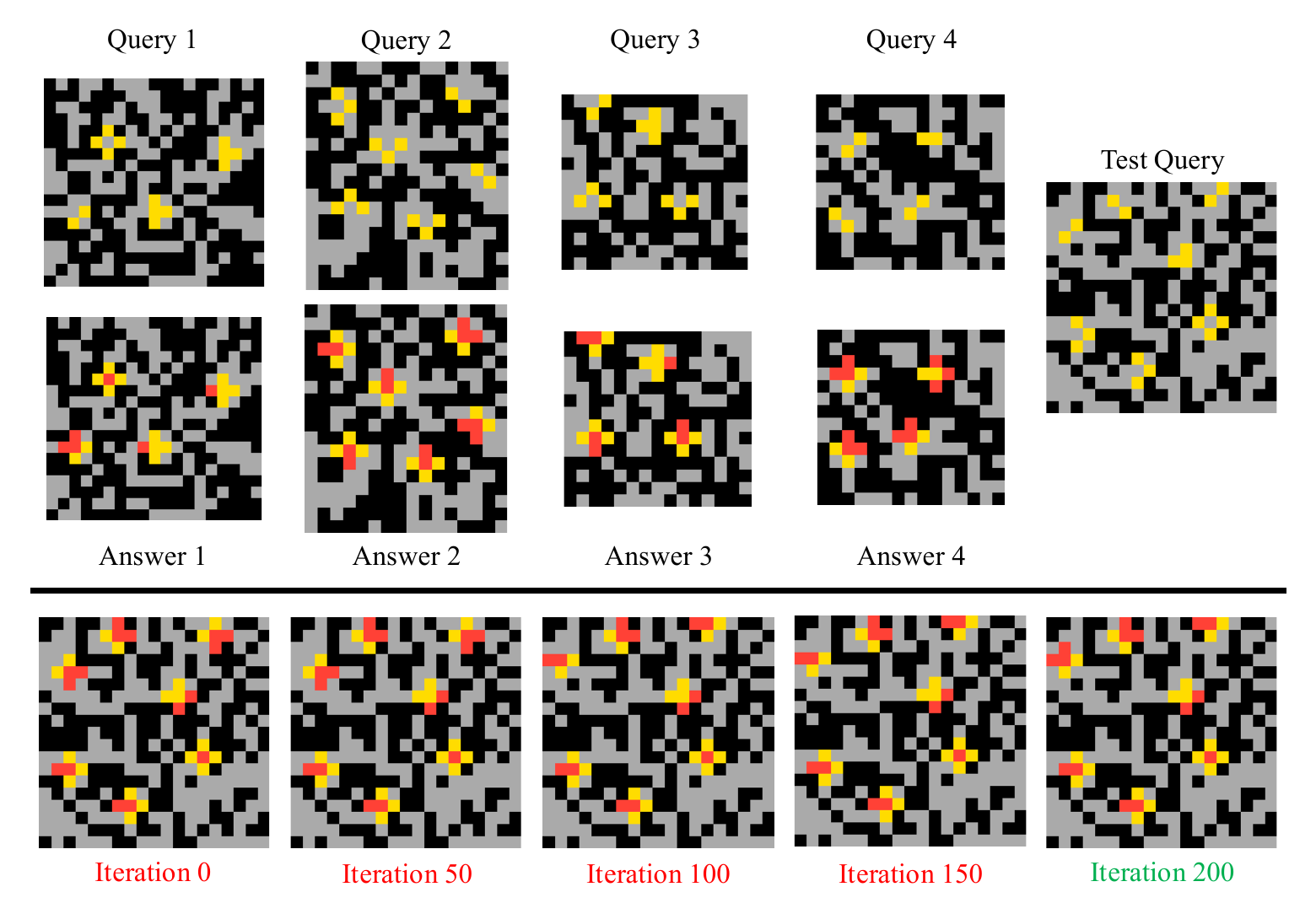}
  \caption{
    \looseness=-10000
    ARC example showing 4 demonstration query-answer pairs, the test query, and LLM predictions at \ourmethodkv{} training iterations 0, 50, 100, 150, and 200. Iteration $0$ is equivalent to ICL, with correct prediction labels in green and incorrect prediction labels in red.
  }
  \label{fig:qualitative}
\end{center}

\section{Conclusion}\label{sec:conclusions}

\looseness=-10000
We introduce \ourmethod{}, a simple and effective method for improving few-shot learning in language models by directly optimizing a prompt or prefix initialized from demonstration tokens. Our method combines the strengths of ICL, which leverages pretrained knowledge by conditioning on task examples at inference time, and prompt-based adaptation, which efficiently adapts to new tasks by tuning a small number of parameters. We develop two versions of this approach: \ourmethodprompt{}, which tunes input-level soft prompts, and \ourmethodkv{}, which tunes layer-specific key and value prefixes derived from the model's activations on demonstration pairs. Across a broad set of benchmarks, both methods outperform ICL, traditional prompt-based tuning, and LoRA variants, with \ourmethodkv{} offering a more favorable tradeoff between performance and efficiency. Through ablation studies, we show that \ourmethodkv{}'s performance depends critically on two design choices: Leave-One-Out Masking and Token Dropout. Additionally, we show that \ourmethodkv{} is robust to varying demonstration count and quality, and scales to large pretrained models.

\looseness=-10000
More broadly, we frame our method within the \ourframework{} framework, which encompasses approaches that leverage in-context demonstration pairs to adapt either the model weights or its context at inference time. Our findings highlight that optimizing the lightweight context, rather than the model, is a powerful and scalable direction for few-shot learning. This approach achieves competitive performance with TTT in significantly less training time. Moreover, we show that \ourmethodkv{} can be applied after TTT to improve performance further, suggesting that context and model adaptation can be effectively combined.

\paragraph{Limitations and Future Work.}
\looseness=-10000
Appendix~\ref{appendix:qualitative_failure} identifies a potential limitation of \ourmethodkv{}: it may be prone to overfitting on certain tasks. Future directions to improve \ourmethodkv{} include exploring stronger regularization techniques beyond Token Dropout, or applying KV cache compression techniques~\citep{l2compression, fastgen, minicache} to compress \ourmethodkv{}'s initialization $\Theta_{\mathrm{CT}}$ before training, further improving overall efficiency.

\section*{Acknowledgments}
\looseness=-10000
This work was supported in part by Visko AI, a Google TPU Award, the NYU-KAIST Award A25-0081-002, and the Institute of Information \& Communications Technology Planning Evaluation (IITP) under grant RS-2024-00469482, funded by the Ministry of Science and ICT (MSIT) of the Republic of Korea in connection with the Global AI Frontier Lab International Collaborative Research.
Computational support was provided by NYU High Performance Computing through its resources, services, and staff expertise.

\bibliographystyle{apalike}
\bibliography{ref}

\clearpage

\appendix

\section*{Appendix}

\section{Time Complexity}\label{sec:timecomplexity}

\looseness=-10000
At each training iteration, an LLM's forward and backward passes are dominated by its self-attention operations. Consider a single attention head of dimension $d_h$. Let $L_Q$ denote the number of query tokens and $L_K$ the number of key (and value) tokens. We form the query matrix $\mathbf{Q}\in\mathbb{R}^{L_Q\times d_h}$ and the key and value matrices $\mathbf{K},\mathbf{V}\in\mathbb{R}^{L_K\times d_h}$, then compute
$
  \mathrm{Attention}\left(
    \mathbf{Q},\mathbf{K},\mathbf{V}
  \right)
  \;=\;
  \mathrm{softmax}\,\left(
    \tfrac{\mathbf{Q}\,\mathbf{K}^\top}{\sqrt {d_h}}
  \right)\,\mathbf{V},
$
\looseness=-10000
whose dominant cost is the matrix multiplication $\mathbf{Q}\,\mathbf{K}^\top$, requiring $O(L_Q\,L_K\,d_h)$ operations per head. Because $d_h$ is a constant for a given model, we omit it in our comparisons below.

\looseness=-10000
Next, let $n$ be the number of tokens in the demonstration pair processed as the current sequence in one leave-one-out training example, and let $p$ be the number of additional context, prompt, or prefix tokens per layer. We analyze how the training time of each method in the \ourframework{} framework scales with $n$ and $p$.

\paragraph{Test-Time Training.}
\looseness=-10000
At each layer of each training iteration, TTT prepends $p$ context tokens to the $n$ tokens in the current training sequence and computes keys and values for all of them, giving $L_Q = n + p$ and $L_K = n + p$ with a per-head cost of
\[
  O\left((n+p)^2\right).
\]

\paragraph{CT-Prompt.}
\looseness=-10000
\ourmethodprompt{} prepends $p$ trainable soft token embeddings to the current training sequence and computes their keys and values, also giving $L_Q = n + p$ and $L_K = n + p$ with a per-head cost of
\[
  O\left((n+p)^2\right).
\]

\paragraph{CT-KV.}
\looseness=-10000
Unlike TTT and \ourmethodprompt{}, \ourmethodkv{} prepends $p$ trainable tokens as past keys and values, so these tokens do not generate queries. This yields $L_Q = n$ and $L_K = n + p$ with a per-head cost of only
\[
  O\left(n\,(n+p)\right).
\]

\paragraph{Time Complexity for $k$ Demonstrations.}
\looseness=-10000
Suppose we have $k$ demonstration pairs, each of length $\ell$ (assuming equal length). In TTT, $n=\ell$ is the length of a demonstration pair and $p=(k-1)\ell$ is the summed length of other demonstration pairs. For \ourmethodprompt{} and \ourmethodkv{}, $n$ and $p$ have the same values as TTT because Leave-One-Out Masking masks out one of the in-context demonstration pairs. Table~\ref{tab:complexity} summarizes the per-head costs in $k$ and $\ell$, showing that both \ourmethodprompt{} and TTT incur quadratic cost in $k$, while \ourmethodkv{}'s cost grows only linearly in $k$. This $k$-fold reduction in self-attention complexity explains \ourmethodkv{}'s faster empirical training speed in Table~\ref{tab:main}.

\begin{table}[h]
  \centering
  \caption{
    \looseness=-10000
    Per-head self-attention time complexity for methods with $k$ demonstration pairs of length $\ell$.
  }
  \begin{tabular}{lccc}
    \toprule
    Method & $L_Q$ & $L_K$ & Per-Head Cost \\
    \midrule
    TTT       & $k\ell$ & $k\ell$ & $O\left((k\ell)^2\right)$ \\
    CT-Prompt & $k\ell$ & $k\ell$ & $O\left((k\ell)^2\right)$ \\
    CT-KV     & $\ell$  & $k\ell$ & $O\left(k\,\ell^2\right)$ \\
    \bottomrule
  \end{tabular}
  \label{tab:complexity}
\end{table}

\section{Prompt Tuning and Prefix Tuning with Other Initialization Schemes}\label{appendix:moreinit}

\begin{center}
  \captionsetup{type=table,hypcap=false}
  \small
  \caption{
    \looseness=-10000
    Ablation of initialization schemes for Prompt Tuning and Prefix Tuning. We show the means of accuracies over 5 seeds with different sets of demonstration pairs per task (except for ARC because it has fixed demonstration pairs).
  }
  \begin{tabular}{@{}lcccc@{}}
      \toprule
      \textbf{Method} & \textbf{NLP-LR} & \textbf{MMLU} & \textbf{BBH} & \textbf{ARC} \\
      \midrule
      Prompt Tuning ($m=32$, uniform) & 39.4  & 34.3 & 34.4  & 5.0 \\
      Prompt Tuning ($m=32$, token)   & 41.4  & 39.2 & 50.8  & 12.0\\
      Prefix Tuning ($m=32$, uniform) & 38.2  & 26.8 & 10.2 & 3.3 \\
      Prefix Tuning ($m=32$, MLP)     & 39.6  & 26.2 & 11.0 & 7.3 \\
      Prefix Tuning ($m=32$, token)   & 42.0  & 39.9 & 52.7  & 9.3 \\
      \bottomrule
  \end{tabular}
  \label{tab:randommlp}
\end{center}

\looseness=-10000
In Table~\ref{tab:main}, we reported Prompt Tuning and Prefix Tuning results using only random-token initialization for their trainable prompts and prefixes. Here, we also follow~\citet{prompttuning} in initializing prompts from a uniform distribution, and~\citet{prefixtuning} in initializing prefixes either from a uniform distribution or from a seed prefix passed through a two-layer MLP (hidden size 512). Table~\ref{tab:randommlp} shows that both Prompt Tuning and Prefix Tuning perform best with random-token initialization, confirming the findings of those works. Therefore, even when compared against these alternative initialization schemes, \ourmethodprompt{} and \ourmethodkv{} continue to deliver superior performance.

\section{More Details on Experiment Setup}\label{sec:hyperparams}

\looseness=-10000
We detail below our hyperparameter settings for the results reported in Table~\ref{tab:main} and Table~\ref{tab:randommlp}. For TTT, we follow~\citet{ttt} by using a LoRA learning rate of $10^{-4}$, sampling a random permutation of the $k$ demonstration pairs at each training step, and setting the LoRA rank to $128$ for ARC and $64$ for all other tasks. In our \tttct{} experiments, we find that using a small number of \ourmethodkv{} training iterations further boosts performance on top of a TTT-adapted model.

\looseness=-10000
For all other experiments, we search over learning rates $3\times10^{-4}$, $10^{-3}$, $3\times10^{-3}$ and Token Dropout rates $0, 0.05, 0.1$. We search training iterations $150, 200, 250, 300$ for NLP-LR and ARC experiments, $15, 20, 25, 30$ for MMLU experiments, and $12, 16, 20, 24$ for BBH experiments. Table~\ref{tab:hyperparams} shows our hyperparameter choices. For fair comparison, hyperparameter sweeps are performed for all methods. For \ourmethodprompt{}, \ourmethodkv{}, and \tttct{}, we use Token Dropout rates of $0.05$ for NLP-LR and $0.1$ for MMLU, BBH, and ARC.

\looseness=-10000
Experiments for NLP-LR are performed on a single RTX 8000, while all other experiments are conducted on a single A100. All experiments use the Adam optimizer, a cosine learning rate scheduler with no warm-up, bfloat16 precision, and up to 32 GB of CPU RAM.

\looseness=-10000
To fairly compare efficiency, we train each method with the largest batch size possible for the GPU used in its experiment. Since TTT, Prompt Tuning (m\,=\,\# demo), and \ourmethodprompt{} use more memory than other methods due to computing larger $\mathbf{Q}\mathbf{K}^\top$ matrices (as shown in our derivation in Section~\ref{sec:timecomplexity}), we limit their batch sizes to $4$ for NLP-LR, MMLU, and ARC, and $5$ for BBH. MMLU and BBH models use gradient checkpointing. For all other methods, we use batch size $16$ for NLP-LR, $8$ for MMLU with gradient accumulation of $2$, $2$ for BBH with gradient accumulation of $5$, and full batch for ARC (depending on each task's number of demonstration pairs). All models, unless noted, do not require gradient checkpointing.

\begin{table}[ht]
    \centering
    \setlength{\tabcolsep}{5pt}
    \small
    \caption{
        \looseness=-10000
        Learning rates (LR) and number of training iterations (\#\,iters) used for each method and benchmark.
    }
    \begin{tabular}{@{}l cc cc cc cc@{}}
    \toprule
         & \multicolumn{2}{c}{\textbf{NLP-LR}} & \multicolumn{2}{c}{\textbf{MMLU}} & \multicolumn{2}{c}{\textbf{BBH}} & \multicolumn{2}{c}{\textbf{ARC}} \\
        \cmidrule(lr){2-3}\cmidrule(lr){4-5}\cmidrule(lr){6-7}\cmidrule(lr){8-9}
        \textbf{Method} & \textbf{LR} & \textbf{\#\,iters} & \textbf{LR} & \textbf{\#\,iters} & \textbf{LR} & \textbf{\#\,iters} & \textbf{LR} & \textbf{\#\,iters} \\
        \midrule
        Prompt Tuning (m\,=\,32, uniform)       & $3\times10^{-3}$ & 200 & $3\times10^{-3}$ &  25 & $10^{-3}$ & 20  & $3\times10^{-3}$ & 250 \\
        Prompt Tuning (m\,=\,32, token)         & $3\times10^{-3}$ & 200 & $10^{-3}$ &  25 & $3\times10^{-3}$ & 16  & $3\times10^{-3}$ & 200 \\
        Prompt Tuning (m\,=\,\#\,demo, token)   & $10^{-3}$ & 250 & $10^{-3}$ &  20 & $3\times10^{-4}$ & 16  & $3\times10^{-3}$ & 200 \\
        Prefix Tuning (m\,=\,32, uniform)       & $3\times10^{-3}$ & 250 & $3\times10^{-3}$ &  25 & $10^{-3}$ & 20  & $3\times10^{-3}$ & 250 \\
        Prefix Tuning (m\,=\,32, MLP)           & $3\times10^{-3}$ & 250 & $10^{-3}$ &  25 & $3\times10^{-3}$ & 20  & $10^{-3}$ & 200 \\
        Prefix Tuning (m\,=\,32, token)         & $3\times10^{-3}$ & 250 & $3\times10^{-3}$ &  25 & $3\times10^{-3}$ & 16  & $3\times10^{-3}$ & 200 \\
        Prefix Tuning (m\,=\,\#\,demo, token)   & $10^{-3}$ & 250 & $3\times10^{-3}$ &  25 & $3\times10^{-3}$ & 16  & $3\times10^{-3}$ & 200 \\
        \ourmethodprompt{}                      & $10^{-3}$ & 250 & $10^{-3}$ &  25 & $3\times10^{-4}$ & 12  & $10^{-3}$ & 250 \\
        \ourmethodkv{}                          & $10^{-3}$ & 200 & $3\times10^{-3}$ &  20 & $10^{-3}$ & 16  & $3\times10^{-3}$ & 200 \\
        TTT                                     & $10^{-4}$ & 250 & $10^{-4}$ &  20 & $10^{-4}$ &  8  & $10^{-4}$ & 200 \\
        \tttct{}                                & $10^{-3}$ &  25 & $10^{-4}$ &   5 & $10^{-3}$ &  8  & $10^{-3}$ &  25 \\
        \bottomrule
    \end{tabular}
    \label{tab:hyperparams}
\end{table}

\section{Parameter-Efficient Variants of CT-KV} \label{sec:paramefficient}

\looseness=-10000
In this section, we explore two variants of \ourmethodkv{} that reduce the number of trainable prefix parameters. We use the notation from Section~\ref{sec:method}.

\paragraph{CT-V.}
\looseness=-10000
We partition the trainable prefix $\Theta_{\mathrm{CT}}$ into its key and value components, $\Theta_K$ and $\Theta_V$. Inspired by~\citet{ewc}, we estimate the importance of each trainable parameter $\Theta_r \in \Theta_{\mathrm{CT}}$ by computing its diagonal Fisher term over the $k$ demonstrations in $\mathcal{D}$:
\begin{equation*}
\hat F_r = \frac{1}{k}\sum_{i=1}^k \left( \nabla_{\Theta_r}\log p_\phi(y_i \bigm| \Theta_{\mathrm{CT}}, x_i) \right)^2.
\end{equation*}
\looseness=-10000
$\hat F_r$ provides a relative estimate of how much a change in each parameter $\Theta_r$ affects the model's ability to solve each demonstration pair, representing its importance during training. By averaging $\hat F_r$ across parameters in $\Theta_K$ and $\Theta_V$, we obtain two scalar estimates, $\hat F_K$ and $\hat F_V$, indicating the relative importance of the trainable keys and values, respectively. Based on our findings in Table~\ref{tab:fishervals}, we conclude that $\hat F_V \gg \hat F_K$ for most tasks, suggesting that values play a more significant role. By freezing $\Theta_K \subset \Theta_{\mathrm{CT}}$ and training only $\Theta_V \subset \Theta_{\mathrm{CT}}$, we arrive at \ourmethodv{}, which reduces the number of trainable parameters in \ourmethodkv{} by exactly half.

\begin{center}
    \small
    \captionsetup{type=table,hypcap=false}
    \caption{
        \looseness=-10000
        Average Fisher information for the trainable key parameters $\Theta_K\subset \Theta_{\mathrm{CT}}$ and value parameters $\Theta_V\subset \Theta_{\mathrm{CT}}$ across 5 random selections of $k$ demonstration pairs over each dataset.
    }
    \begin{tabular}{lcc}
    \toprule
        Dataset & \(\hat F_K\) & \(\hat F_V\) \\
        \midrule
        ARC   & \(2.43\times 10^{-9}\) & \(1.02\times 10^{-7}\) \\
        BBH   & \(1.89\times 10^{-6}\) & \(3.99\times 10^{-4}\) \\
        NLP-LR & \(1.44\times 10^{-8}\) & \(8.32\times 10^{-8}\) \\
        MMLU  & \(2.81\times 10^{-8}\) & \(1.42\times 10^{-6}\) \\
        \bottomrule
    \end{tabular}
    \label{tab:fishervals}
\end{center}

\paragraph{CT-Prefix.}
\looseness=-10000
We freeze $\Theta_{\mathrm{CT}}$, average the parameters across tokens to obtain an average prefix $\bar \Theta_{\mathrm{CT}}$, and then form a new trainable $m$-token prefix $\Theta_{\mathrm{prefix}}$ by adding small Gaussian perturbations:
\[
    \Theta_{\mathrm{prefix},i} = \bar\Theta_{\mathrm{CT}} + \epsilon_i,\quad i=1,\ldots,m,
\]
\looseness=-10000
where $\epsilon_i \sim \mathcal{N}(0, 0.02^2 I)$. The model additionally conditions on $\Theta_{\mathrm{prefix}}$, analogous to Prefix Tuning. Since we only train $\Theta_{\mathrm{prefix}}$, this variant has the same number of trainable parameters as Prefix Tuning with $m$ tokens.

\looseness=-10000
We evaluate \ourmethodv{} and \ourmethodprefix{} across all benchmarks and compare them to \ourmethodkv{} in Table~\ref{tab:parameterefficient}, showing that both parameter-efficient variants retain most of the performance gain of \ourmethodkv{} and outperform Prefix Tuning. For \ourmethodv{}, we use the same hyperparameters as \ourmethodkv{} from Section~\ref{sec:hyperparams}. For \ourmethodprefix{}, we find that higher learning rates, $10^{-1}$ for NLP-LR and $5\times10^{-2}$ for other datasets, are needed for better performance.

\begin{table}[h]
    \centering
    \small
    \caption{
        \looseness=-10000
        Accuracies (\%) of \ourmethodkv{}, its parameter-efficient variants, and Prefix Tuning across benchmarks, averaged over 5 seeds (except for ARC because it has fixed demonstration pairs).
    }
    \begin{tabular}{@{}lcccc@{}}
        \toprule
        \textbf{Method}       & \textbf{NLP-LR} & \textbf{MMLU} & \textbf{BBH} & \textbf{ARC} \\
        \midrule
        Prefix Tuning (m\,=\,32) & 42.0 & 39.9 & 52.7 & 9.3 \\
        \ourmethodprefix{}    & 44.0            & 42.6          & 55.9         & 22.8         \\
        \ourmethodv{}         & 44.0            & 43.5          & 57.5         & 23.5         \\
        \ourmethodkv{}        & 44.2            & 43.7          & 57.9         & 23.8         \\
        \bottomrule
    \end{tabular}
    \label{tab:parameterefficient}
\end{table}

\section{Number of Trainable Parameters}\label{appendix:numparams}

\looseness=-10000
Corresponding to the performance shown in Table~\ref{tab:main}, we report the average number of trainable parameters for each method across tasks in Table~\ref{tab:nparams}. Note that although \ourmethodkv{}'s number of trainable parameters scales with the number of demonstration tokens, it still trains significantly fewer parameters on average per task than the number of LoRA parameters used by TTT. Following~\citet{ttt}, we use task instructions for BBH and set the instruction prompt or prefix to be trainable as well. We omit Zero-Shot and ICL from this comparison because they do not involve any trainable parameters.

\begin{table}[h]
    \centering
    \small
    \caption{
        \looseness=-10000
        Number of trainable parameters (in thousands) for each method across benchmarks, corresponding to entries in Table~\ref{tab:main}.
    }
    \begin{tabular}{@{}lcccc@{}}
        \toprule
        \textbf{Method} & \textbf{NLP-LR} & \textbf{MMLU} & \textbf{BBH} & \textbf{ARC} \\
        \midrule
        Prompt Tuning (m\,=\,32)                      &   41    &   98    &   229    &    66    \\
        Prompt Tuning (m\,=\,\#\,demo)        &  578    & 2160  & 3656   & 2743   \\
        Prefix Tuning (m\,=\,32)                      & 2949  & 1835  & 3668   &   524    \\
        Prefix Tuning (m\,=\,\#\,demo)        & 41634 & 40327 & 58501  & 21944  \\
        TTT                        & 47186 & 89915 &157286  & 84935  \\
        \ourmethodprompt{}                        &  578    & 2160  & 3656   & 2743   \\
        \ourmethodprefix{}                        &2949   &1835   &3668    &   524    \\
        \ourmethodv{}                              &20817  &20163  &29250   &10972   \\
        \ourmethodkv{}                            &41634  &40327  &58501   &21944   \\
        \tttct{}                                  &88820  &130242 &215787  &106878  \\
        \bottomrule
    \end{tabular}
    \label{tab:nparams}
\end{table}

\section{Memory Profiling}

\looseness=-10000
We compare the training-time memory usage of \ourmethodkv{} and our baselines in Table~\ref{tab:training-cost}. Values indicate the peak GPU memory usage (MB) when training with batch size 1 for one epoch with no gradient accumulation or checkpointing. Results are averaged across all tasks of each benchmark. We omit Zero-Shot and ICL because they do not require training. Since \tttct{} simply runs the two methods sequentially, its memory usage would be the maximum of TTT and \ourmethodkv{}. All other experimental settings remain identical to those in Table~\ref{tab:main}.

\begin{table}[h]
    \centering
    \setlength{\tabcolsep}{6pt}
    \small
    \caption{
        \looseness=-10000
        Memory profiling of \ourmethodkv{} and baselines.
    }
    \begin{tabular}{lcccc}
        \toprule
        \textbf{Method} & \textbf{NLP-LR} & \textbf{MMLU} & \textbf{BBH} & \textbf{ARC} \\
        \midrule
        Prompt Tuning (m\,=\,32)      & 3601 & 13051 & 31640 & 4486 \\
        Prompt Tuning (m\,=\,\#\,demo) & 7319 & 18868 & 37493 & 6451 \\
        Prefix Tuning (m\,=\,32)      & 3588 & 13220 & 31549 & 4598 \\
        Prefix Tuning (m\,=\,\#\,demo) & 5648 & 15570 & 33292 & 5340 \\
        TTT                        & 8883 & 18690 & 54752 & 7232 \\
        \ourmethodprompt{}                  & 7318 & 18911 & 37571 & 6629 \\
        \ourmethodkv{}                      & 5648 & 15569 & 33292 & 5340 \\
        \bottomrule
    \end{tabular}
    \label{tab:training-cost}
\end{table}

\looseness=-10000
Together with the accuracy results in Table~\ref{tab:main}, our memory profiling shows that \ourmethodkv{} uses less memory than TTT, outperforms Prefix Tuning (m\,=\,\#\,demo) at the same memory usage, and incurs less than $20\%$ relative memory overhead compared to Prefix Tuning (m\,=\,32) on MMLU, BBH, and ARC, while achieving significantly higher accuracy.

\section{Many-Shot Performance on BBH}\label{appendix:manyshot}

\looseness=-10000
We extend the demonstration-count ablation in Section~\ref{sec:robustness} to the many-shot regime on BBH using Llama 3-8B-Instruct, restricted to the $8$ tasks whose $100$-shot prompts fit within the $8$K-token context window. Table~\ref{tab:manyshot} reports these results. We use the same hyperparameters selected for the BBH results in Table~\ref{tab:main}.

\begin{table}[h]
    \centering
    \setlength{\tabcolsep}{6pt}
    \small
    \caption{
        \looseness=-10000
        Many-shot performance of ICL and \ourmethodkv{} on BBH with Llama 3-8B-Instruct.
    }
    \begin{tabular}{l c c c c}
        \toprule
        \textbf{Method} & \textbf{Epochs} & \textbf{25-shot} & \textbf{50-shot} & \textbf{100-shot} \\
        \midrule
        ICL             & N/A       & 48.7          & 46.8          & 46.5          \\
        \midrule
        \multirow{5}{*}{\ourmethodkv{}}
                        & 3         & 50.5          & 51.6          & 51.2          \\
                        & 10        & 52.6          & 56.1          & 57.5          \\
                        & 20        & 54.0          & 58.6          & 61.1          \\
                        & 40        & 54.9          & $\mathbf{59.8}$ & $\mathbf{62.6}$ \\
                        & 60        & $\mathbf{55.3}$ & 59.0          & 62.0          \\
        \bottomrule
    \end{tabular}
    \label{tab:manyshot}
\end{table}

\looseness=-10000
ICL degrades from $48.7\%$ at $25$-shot to $46.5\%$ at $100$-shot, while \ourmethodkv{} keeps improving as demonstrations are added, widening its absolute accuracy gap over ICL from $6.6\%$ to $16.1\%$ by $100$-shot.

\section{Stability Under Extended Training}\label{appendix:overfit}

\looseness=-10000
We sweep the number of training epochs on BBH using the Table~\ref{tab:main} setup to compare \ourmethodkv{} against Prefix Tuning under prolonged optimization. Table~\ref{tab:overfit} reports these results. The ICL baseline accuracy on BBH is $50.44\%$. \ourmethodkv{}'s accuracy is essentially flat from epoch $16$ to $64$ ($57.91 \to 57.88$). Prefix Tuning peaks at epoch $32$ ($52.73$) and declines by epoch $64$ ($50.49$), falling back to the ICL baseline despite tuning fewer parameters than \ourmethodkv{}.

\begin{table}[h]
    \centering
    \setlength{\tabcolsep}{6pt}
    \small
    \caption{
        \looseness=-10000
        BBH accuracy of \ourmethodkv{} and Prefix Tuning as the number of training epochs increases.
    }
    \begin{tabular}{l c c c c c}
        \toprule
        \textbf{Epochs} & 4 & 8 & 16 & 32 & 64 \\
        \midrule
        Prefix Tuning   & 42.15 & 49.35 & 51.00 & $\mathbf{52.73}$ & 50.49 \\
        \ourmethodkv{}  & 53.13 & 55.46 & 57.91 & $\mathbf{57.93}$ & 57.88 \\
        \bottomrule
    \end{tabular}
    \label{tab:overfit}
\end{table}

\section{Additional Qualitative Samples vs. Training Iteration}\label{appendix:qualitative_iterations}

\makeatletter
\setlength{\@fptop}{0pt}
\makeatother

\begin{figure}[t]
  \centering
  \includegraphics[trim={10.55cm 20.53cm 11.04cm 0.39cm},clip,width=0.9\textwidth]{assets/supp_arc.pdf}
  \caption{
    \looseness=-10000
    Additional ARC example showing 4 demonstration query-answer pairs, the test query, and LLM predictions at \ourmethodkv{} training iterations 0, 50, 100, 150, and 200. Iteration $0$ is equivalent to ICL, with correct prediction labels in green and incorrect prediction labels in red.
  }
  \label{fig:supparc}
\end{figure}

\looseness=-10000
Figure~\ref{fig:supparc} shows a second ARC task, complementing the main qualitative example in Figure~\ref{fig:qualitative}. At iteration $0$, the model shows a strong bias toward filling orange squares with yellow. As \ourmethodkv{} training progresses, the model gradually learns to fill each orange square with the correct color.

\begin{figure}[t]
  \centering
  \includegraphics[trim={15pt 35pt 15pt 65pt},clip,width=1.0\textwidth]{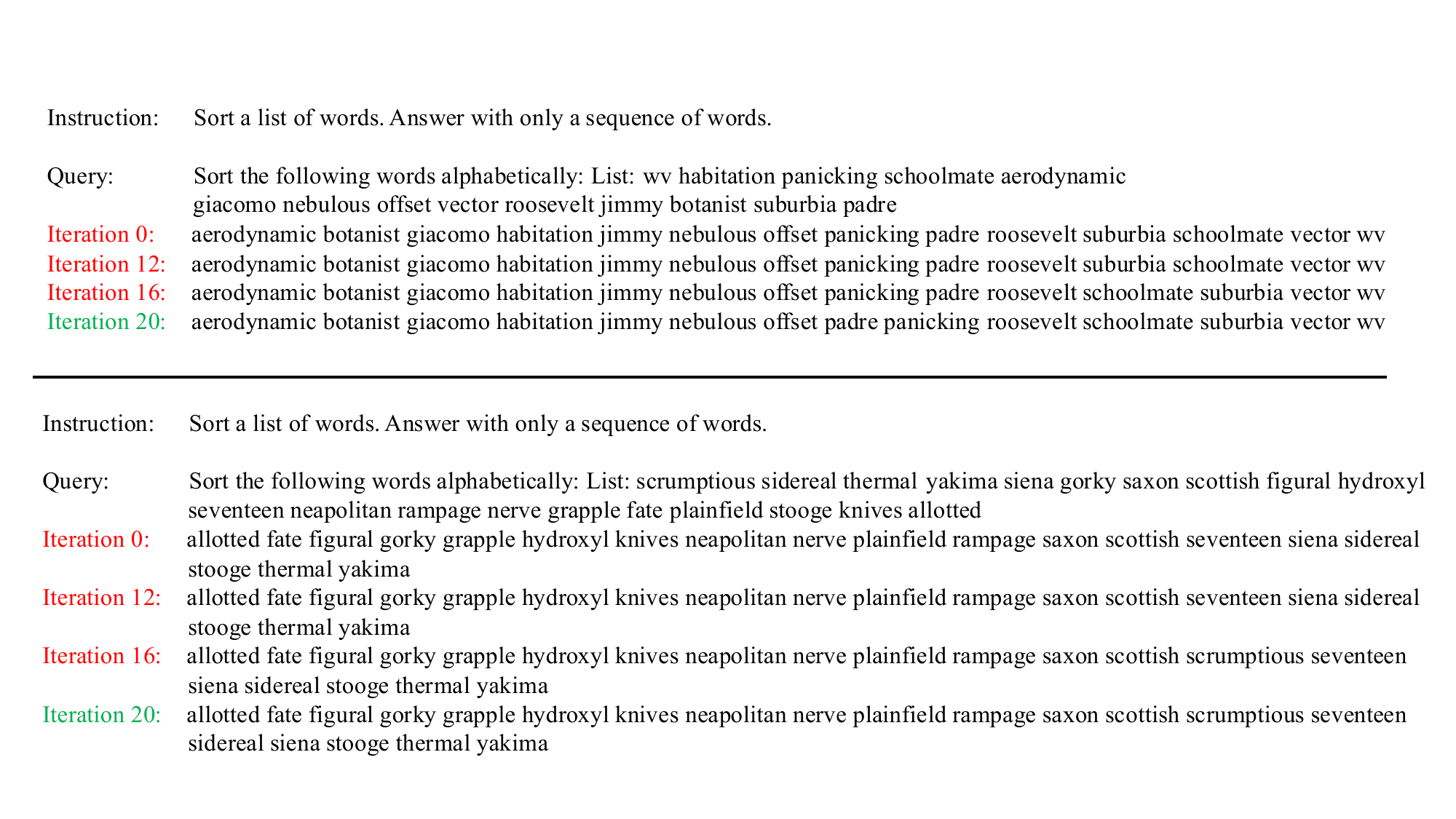}
  \caption{
    \looseness=-10000
    We display LLM predictions at \ourmethodkv{} training iterations 0, 12, 16, and 20 for two queries from the task ``word sorting'' in BBH. We omit the $16$ demonstration pairs provided with each query for brevity. We color-code correct prediction labels in green and incorrect prediction labels in red.
  }
  \label{fig:suppbbh}
\end{figure}

\looseness=-10000
Similarly, for BBH, in Figure~\ref{fig:suppbbh}'s top query, the model initially predicts ``padre, panicking'' and ``schoolmate, suburbia'' in reversed order at iteration $0$. During \ourmethodkv{} training, the model learns to compare later letters for sorting and eventually answers the query correctly. Likewise, for the bottom query, \ourmethodkv{} helps the model avoid omitting the word ``scrumptious'' from its outputs and sort the words ``sidereal, siena'' into the correct order by comparing later letters.

\clearpage

\section{Additional ARC Qualitative Comparison}\label{appendix:qualitative_failure}

\looseness=-10000
We compare our \ourmethodkv{} to ICL on the 400 ARC evaluation tasks. Figure~\ref{fig:qualitative_failure} shows one failure case for each method, where the other successfully solves the task. In the left task, \ourmethodkv{} adapts to the demonstration pairs and solves a geometric puzzle that crops the upper-left portion of objects in the query. In the right task, \ourmethodkv{} predicts the wrong output. Two of the three demonstration answers on the right have 3-row by 4-column grids, which may bias \ourmethodkv{} toward predicting the same shape during optimization.

\begin{center}
  \captionsetup{type=figure,hypcap=false}
  \includegraphics[page=2,trim={0.5cm 8cm 0.5cm 7.8cm},clip,width=1.0\textwidth]{assets/qualitative.pdf}
  \caption{
    \looseness=-10000
    Left is an ARC task that \ourmethodkv{} successfully solves, but ICL does not. Right is an ARC task that ICL solves, but \ourmethodkv{} does not.
  }
  \label{fig:qualitative_failure}
\end{center}

\clearpage

\section{Demonstration Pairs for Figure~\ref{fig:dataset}}\label{appendix:qualitative}

\begin{figure}[h]
  \centering
  \includegraphics[trim={0cm 9.5cm 0cm 0cm},clip,width=1.0\textwidth]{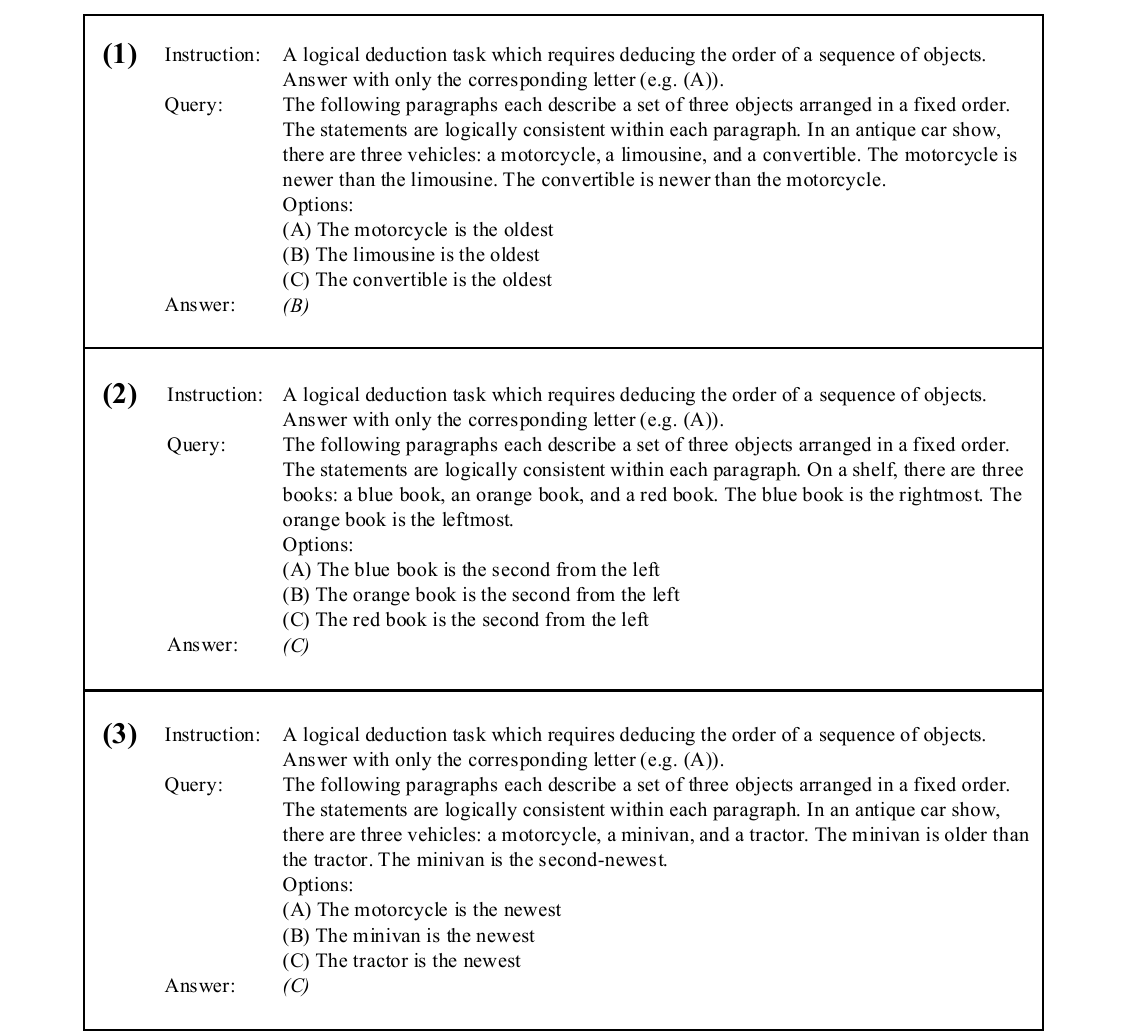}
  \caption{
    \looseness=-10000
    Three demonstration pairs for the BBH task from Figure~\ref{fig:dataset}.
  }
  \label{fig:bbhdemo}
\end{figure}

\begin{figure}[h]
  \centering
  \includegraphics[trim={0cm 20.1cm 0cm 0cm},clip,width=1.0\textwidth]{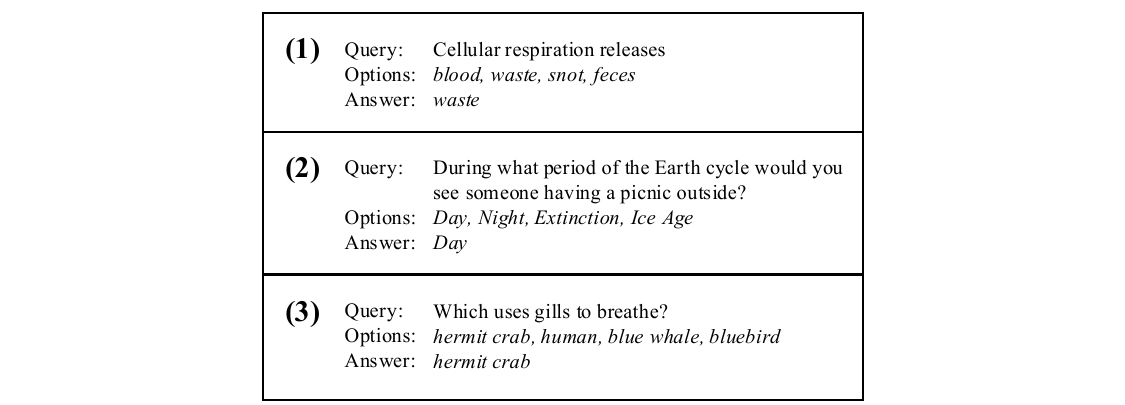}
  \caption{
    \looseness=-10000
    Three demonstration pairs for the NLP-LR task from Figure~\ref{fig:dataset}.
  }
  \label{fig:nlpdemo}
\end{figure}

\begin{figure}[h]
  \centering
  \includegraphics[trim={0cm 20cm 0cm 0cm},clip,width=1.0\textwidth]{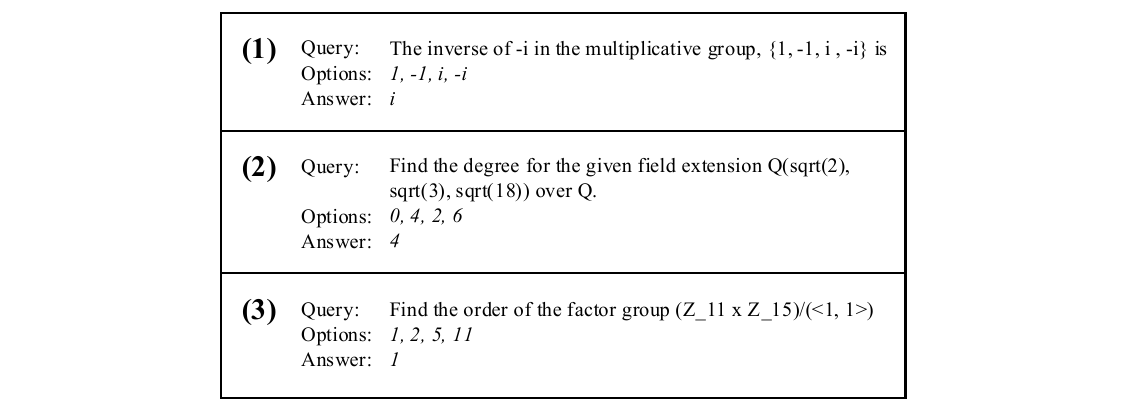}
  \caption{
    \looseness=-10000
    Three demonstration pairs for the MMLU task from Figure~\ref{fig:dataset}.
  }
  \label{fig:mmludemo}
\end{figure}

\looseness=-10000
We present three demonstration pairs from each of BBH, NLP-LR, and MMLU in Figure~\ref{fig:bbhdemo}, Figure~\ref{fig:nlpdemo}, and Figure~\ref{fig:mmludemo}, respectively.

\end{document}